\documentclass[10pt, a4paper]{article}

\usepackage[final]{lrec2026} 

\usepackage{comment}
\usepackage{siunitx}
\usepackage{bm}

\usepackage{booktabs}
\usepackage[ruled,vlined]{algorithm2e}
\usepackage{xcolor}
\usepackage{soul}
\usepackage{multirow}
\usepackage{cleveref}
\definecolor{pastelblue}{HTML}{AEC6CF}

\definecolor{pink}{HTML}{FF8FD0}

\title{Can LLMs Evaluate What They Cannot Annotate? Revisiting LLM Reliability in Hate Speech Detection}

\name{Paloma Piot\textsuperscript{1}, David Otero\textsuperscript{1}, Patricia Martín-Rodilla\textsuperscript{2}, Javier Parapar\textsuperscript{1}}

\address{
\textsuperscript{1}IRLab, CITIC Research Centre, Universidade da Coruña, Spain \\
\{paloma.piot, david.otero.freijeiro, javier.parapar\}@udc.es \\
\textsuperscript{2}IEGPS CSIC, Santiago de Compostela, Spain \\
p.m.rodilla@iegps.csic.es
}

\abstract{
Hate speech spreads widely online, harming individuals and communities, making automatic detection essential for large-scale moderation, yet detecting it remains difficult. Part of the challenge lies in subjectivity: what one person flags as hate speech, another may see as benign. Traditional annotation agreement metrics, such as Cohen's $\kappa$, oversimplify this disagreement, treating it as an error rather than meaningful diversity. Meanwhile, Large Language Models (LLMs) promise scalable annotation, but prior studies demonstrate that they cannot fully replace human judgement, especially in subjective tasks. In this work, we reexamine LLM reliability using a subjectivity-aware framework, cross-Rater Reliability (xRR), revealing that even under fairer lens, LLMs still diverge from humans. Yet this limitation opens an opportunity: we find that LLM-generated annotations can reliably reflect performance trends across classification models, correlating with human evaluations. We test this by examining whether LLM-generated annotations preserve the relative ordering of model performance derived from human evaluation (i.e. whether models ranked as more reliable by human annotators preserve the same order when evaluated with LLM-generated labels). Our results show that, although LLMs differ from humans at the instance level, they reproduce similar ranking and classification patterns, suggesting their potential as proxy evaluators. While not a substitute for human annotators, they might serve as a scalable proxy for evaluation in subjective NLP tasks. \\
\newline 
}
\begin{document}

\maketitleabstract

\section{Introduction}

Hate speech, defined as ``\textit{language characterised by offensive, derogatory, humiliating, or insulting discourse~\cite{fountahatespeech} that promotes violence, discrimination, or hostility towards individuals or groups~\cite{davidsonhatespeech} based on attributes such as race, religion, ethnicity, or gender~\cite{hatelingo2018elsherief,ElSherief2018,hatemm2023}}'', continues to pose a major challenge across online platforms, undermining both individual safety and social cohesion~\cite{GonzlezBailn2022}. Recent studies estimate that nearly one in three young people have experienced some form of cyberbullying~\cite{KansokDusche2022}, while around half of Black or African American adults report having faced racial harassment online~\cite{ADL2024}. This growing prevalence of hate speech in social platforms underscores the urgent need for reliable automatic detection methods.

Despite notable progress in hate speech detection ~\cite{glavas-etal-2020-xhate,wang2022toxicitydetectiongenerativepromptbased,plaza-del-arco-etal-2023-respectful,roy-etal-2023-probing,Piot2025}, the quality and consistency of human-annotated data remain central challenges~\cite{toraman-etal-2022-large,Ljubei2022,barbarestani-etal-2024-content}. The task is inherently subjective~\cite{10.1145/3308560.3317083,kanclerz-etal-2022-ground,Hettiachchi2023}, as judgements often depend on cultural norms, speaker intent, and contextual interpretation~\cite{rottger-etal-2022-two,davani-etal-2023-hate,dehghan2025dealingannotatordisagreementhate}. As a result, annotation disagreement is not always an error but may reflect legitimate diversity in interpretation~\cite{10.1007/978-3-031-08974-9_54,davani-etal-2022-dealing,barbarestani-etal-2024-content}.
Conventional inter-annotator agreement (IAA) measures, such as Cohen's $\kappa$, assume a single objective ground truth and penalise all disagreement equally~\cite{doi:10.1177/001316446002000104}, making them improper for tasks involving subjectivity and ambiguity such as hate speech detection~\cite{marchal-etal-2022-establishing,plank-2022-problem,bassi-etal-2025-annotating}.

To address this limitation, subjectivity-aware metrics have been proposed. Among them, the xRR (cross-Rater Reliability) framework~\cite{Wong2021} has emerged as a promising alternative, as it evaluates how consistently annotators reproduce each other's labelling behaviour rather than enforcing a single gold label. This shift better captures pluralism in human judgements, offering a more realistic picture of annotation quality in socially grounded tasks~\cite{dutta2023modelingsubjectivitybymimicking,prabhakaran-etal-2024-grasp}.

Meanwhile, LLMs have been explored as potential automatic annotators, offering scalability and cost-efficiency~\cite{doi:10.1073/pnas.2305016120,doi:10.1177/08944393241286471}. However, current evidence shows that LLMs cannot yet replace human annotators, particularly in tasks that require social or contextual interpretation~\cite{pmlr-v239-mohta23a,tseng2025evaluatinglargelanguagemodels,baumann2025largelanguagemodelhacking}. Despite these advances, most prior studies assessing LLM annotations rely on traditional agreement metrics~\cite{Matter2024,Giorgi2025}, which may undervalue their performance under subjective conditions. This raises an important motivational question: does evaluating LLMs-based annotations with subjectivity-aware metrics alter the conclusions we may draw about their reliability as annotators?

Building on this motivation, we first investigate the role of LLMs as annotators in multilingual hate speech detection (English and Spanish). Our findings show that, even when adopting xRR to account for subjectivity, LLMs still fail to capture the full spectrum of human annotation patterns. Nonetheless, a complementary role emerges: rather than replacing human annotators, LLMs may serve as system evaluators. We conduct a second experiment to test whether LLM-generated annotations can reproduce the relative ranking of hate speech detection models established by human annotators. This framework reveals that LLMs, despite instance-level divergence, can reliably approximate classification models performance trends, offering a scalable path for comparative evaluation without requiring full human annotation.

Our study is structured around three research questions that build sequentially on one another:

\begin{itemize}
    \item \textit{\textbf{RQ1}: How do traditional agreement metrics capture annotation quality in hate speech detection, both for human and LLM-generated labels?}
    \item \textit{\textbf{RQ2}: If using subjectivity-aware metrics, do LLMs demonstrate greater reliability as annotators, or do they still diverge from human annotation patterns?}
    \item \textit{\textbf{RQ3}: Even if LLMs are not reliable annotators, can their annotation patterns be leveraged for evaluation, preserving the relative ordering of hate speech detection models observed under human judgements?}
\end{itemize}

We structure our study around two complementary experiments: $(i)$ an agreement analysis, addressing \textit{\textbf{RQ1}} and \textit{\textbf{RQ2}}, where we compare traditional and subjectivity-aware metrics to assess annotation reliability across human and LLM judgements; and $(ii)$ a ranking correlation experiment, addressing \textit{\textbf{RQ3}}, where we test whether LLM-based annotations maintain the relative ordering of model performance derived from human evaluation (for example, whether models ranked as more reliable by human annotators preserve the same ordering when evaluated with LLM-generated labels). Through these questions, we aim to deepen the understanding of how metric selection choice shapes annotation assessment, and to clarify the emerging role of LLMs as evaluators in socially grounded tasks.

\section{Related Work}

\paragraph{Hate speech detection.} Hate speech detection remains a challenging task due to its inherently subjective nature, where individual perceptions, social-cultural context and linguistic nuance often lead to divergent annotations~\cite{10.1145/3232676,kanclerz-etal-2022-ground}. IAA in this domain tends to be low, reflecting the ambiguity and contextual dependency of what constitutes hate speech~\cite{toraman-etal-2022-large,Ljubei2022,barbarestani-etal-2024-content}. Prior research has emphasised that such disagreement is not merely noise, but a signal of underlying subjectivity in annotation~\cite{10.1007/978-3-030-77091-4_26,leonardelli-etal-2021-agreeing,guellil-etal-2024-annotators}. This challenge has prompted renewed attention to how annotation quality is measured, especially in socially grounded tasks where disagreement among annotators is common.

\paragraph{IAA in subjective tasks.} Traditional metrics such as Cohen's $\kappa$~\cite{doi:10.1177/001316446002000104} and Fleiss's $\kappa$~\cite{Shrout1979} are commonly used to assess annotation consistency. However, they often underestimate reliability in socially grounded tasks, as they penalise legitimate subjective variation~\cite{pavlick-kwiatkowski-2019-inherent}. Recent work has proposed ways to better capture human uncertainty, including soft-label modelling~\cite{10.1613/jair.1.12752} and multi-annotator models, where each annotator's judgements are treated as separate subtasks while sharing a common learned representation~\cite{davani-etal-2022-dealing}. Yet, less attention has been paid to how agreement itself is measured. Metrics such as xRR~\cite{Wong2021} explicitly address subjectivity in annotation, capturing consistent patterns of disagreement rather than treating them as noise or error. For instance, in subjective annotation tasks, \citet{dutta2023modelingsubjectivitybymimicking} demonstrate that xRR reveals community-specific perspectives in toxic comment annotation, while \citet{prabhakaran-etal-2024-grasp} leverage xRR to analyse inter-group disagreement patterns, highlighting how perspectives vary across demographic groups.

\paragraph{LLMs as annotators.} LLMs have been investigated as tools to assist or replace human annotators, providing a scalable way to generate labels for large datasets~\cite{doi:10.1073/pnas.2305016120,doi:10.1177/08944393241286471}. Studies show that while LLMs can approximate human judgements in certain tasks, their reliability is generally lower than humans, especially in tasks that involve social nuance or context-dependent interpreation~\cite{pmlr-v239-mohta23a,tseng2025evaluatinglargelanguagemodels,baumann2025largelanguagemodelhacking}. Recent work has therefore focused on understanding where LLMs succeed or fail in mimicking human annotations, leaving open questions about alternative uses.

Given this, our study is motivated by the gap in understanding LLM performance under subjectivity-aware metrics and its implications for evaluation. We focus on where LLMs diverge from humans and how they might serve as model evaluators in subjective tasks like hate speech.

\section{Experimental Setup}

To address our three research questions, we design a two-stage experimental study grounded in a unified setup. All experiments rely on the same hate speech datasets, language models, and LLM-generated annotations. This section details the datasets, the selected models, the prompting and annotation procedure, and the metrics used to assess IAA.

\subsection{Data}

For this study, we build upon the \textsc{MetaHate}~\cite{Piot2024} and \textsc{MetaHateES}~\cite{piot2025bridginggapshatespeech} collections, two large scale meta-collections that consolidate the major publicly available hate speech datasets in English and Spanish, respectively. Both resources were designed to standardise and unify heterogeneous hate speech resources. \textsc{MetaHate} integrates \num{36} publicly available datasets in English covering diverse social media platforms, domains and annotation schemes, while \textsc{MetaHateES} includes \num{10} different sources, focusing on European Spanish hate speech. We focus on these languages as high-resource languages enabling controlled cross-lingual comparison~\cite{10.1145/3232676,Poletto2020,Vidgen2020}, and providing robust baselines for future extension to low-resource varieties.

Given the focus of this work on annotation subjectivity and inter-annotator reliability, we selected from them only those datasets that contained individual-level annotation, that could enable a fine-grained comparison between human and LLM judgements. Following this criterion, we initially identified two datasets from \textsc{MetaHate}: \textsc{HateXplain}~\cite{mathewb2020hatexplain} and Measuring Hate Speech (\textsc{MHS})~\cite{kennedy2020constructingintervalvariablesfaceted,sachdeva-etal-2022-measuring}; and two from \textsc{MetaHateES}: \textsc{DETESTS}~\cite{detests2024} and \textsc{EXIST}~\cite{exist2022}. Table~\ref{tab:potential-datasets} summarises these datasets. 

\begin{table}[h]
    \centering
    \begin{tabular}{llcr}
    \toprule
    \textbf{Dataset} & \textbf{Lang.} & \textbf{\# annotations} & \textbf{\# posts} \\
    \midrule
    EXIST & ES & \num{6} & \num{4209} \\
    DETESTS & ES &  \num{3} & \num{9906} \\
    HateXplain & EN &  \num{3} & \num{20109} \\ 
    MHS & EN & \num{1}-\num{815} & \num{39565} \\
    \bottomrule
    \end{tabular}
    \caption{Overview of the \emph{potential} datasets that can be used in our experiments. ``Lang.'' denotes the dataset language (ES $=$ Spanish, EN $=$ English). ``\# annotations'' indicates the number of independent judgements available per post, and ``\# posts'' reports the number of unique, non-duplicated instances.}
    \label{tab:potential-datasets}
\end{table}

The task under study is binary hate speech detection, where each post is classified as either \texttt{hate} or \texttt{non-hate}. Most selected datasets follow this formulation directly. However, some require label harmonisation to ensure cross-dataset consistency. \textsc{HateXplain} was originally annotated with three labels: ``normal'', ``offensive'' and ``hate speech''. Following prior work~\cite{Piot2024}, we merged ``offensive'' category into \texttt{no-hate}, as offensive content does not fit our hate speech definition. The \textsc{MHS} dataset includes an additional label, ``unclear''; we removed all instances with at least one ``unclear'' annotation to maintain consistency with the binary setup. 

To ensure a fair comparison across datasets, we restricted our analysis to instances with exactly three human annotators. Therefore, the \textsc{EXIST} dataset was excluded, and we only retained \textsc{MHS} instances meeting this criterion. After unifying label spaces, we derived a gold label for each post based on the majority vote across annotators, and recorded whether the gold label was unanimous or not. Table~\ref{tab:datasets} summarises the final datasets and the number of posts used in our experiments.

\begin{table}[h]
    \centering
    \resizebox{\columnwidth}{!}{%
    \begin{tabular}{llc rrr}
    \toprule
    \textbf{Dataset} & \textbf{Lang.} & \textbf{\# annot.} & \textbf{\# posts} & \textbf{\# unanimous} & \textbf{\% hate} \\
    \midrule
    DETESTS & ES &  \num{3} & \num{9906} & \num{8589} & \num{26.30} \\
    HateXplain & EN &  \num{3} & \num{20109} & \num{13910} & \num{29.49} \\
    MHS & EN & \num{3} & \num{9133} & \num{6428} & \num{25.64} \\
    \bottomrule
    \end{tabular}
    }
    \caption{Overview of the datasets \emph{included} in our experiments. ``Lang.'' denotes the dataset language (ES $=$ Spanish, EN $=$ English). ``\# annot.'' indicates the number of independent judgements used per post, ``\# posts'' reports the number of unique, non-duplicated instances, and ``\# unanimous'' reports the number of posts with unanimous labels.}
    \label{tab:datasets}
\end{table}


\subsection{Models}

To generate LLM-based annotations for our experiments, we selected three state-of-the-art instruction-tuned models, all running in 4-bit quantized versions to optimise computational efficiency without compromising performance. The choice of three models mirrors the number of human annotators per dataset, ensuring a comparable annotation setup. Below are the details of the LLMs we employed:

\begin{itemize}
    \item \texttt{Llama 3.1} (\texttt{Llama-3.1-8B-Instruct}) \cite{grattafiori2024llama3herdmodels}: Developed by Meta AI, this model is part of the Llama 3 series, which includes models ranging from 8B to 405B parameters. The 8B variant is optimised for instruction-following tasks and has demonstrated strong performance across multilingual and reasoning benchmarks.
    \item \texttt{Nemo} (\texttt{Mistral-Nemo-Instruct-2407}) \cite{mistralnemo}: This model comes from a collaboration between Mistral AI and NVIDIA. It is an instruction-tuned model based on the Mistral architecture and it is designed to excel in multilingual tasks and has been fine-tuned to improve reasoning and coding accuracy.
    \item \texttt{DeepSeek} (\texttt{DeepSeek-R1-Distill-Llama-8B})~\cite{deepseekai2025deepseekr1incentivizingreasoningcapability}: This model is a distilled version of the \texttt{Llama-3.1-8B-Instruct} model, fine-tuned using samples from DeepSeek-R1. The distillation process aims to retain the performance of the original model while reducing computational requirements. It is particularly noted for its efficiency and effectiveness in reasoning tasks.
\end{itemize}

\subsection{LLM judgements}

To obtain automatic annotations, we queried each selected LLM using a prompt template. The models were asked to assess whether a given social media message contained hate speech and respond strictly with a binary label as: \texttt{True} (\texttt{hate}) or \texttt{False} (\texttt{no-hate}). No explicit definition or examples of hate speech were provided; the models generated labels based solely on their internal understanding of the task.

\paragraph{Prompts.} All models were queried with the same two-part prompt to keep the setup consistent. The system prompt introduced the task as a moderation scenario, and the base instruction told the model to return a binary decision. Tables~\ref{tab:system-prompt} and~\ref{tab:base-instr} provide the exact wording.

\begin{table}[ht!]
    \tabcolsep=0.08cm
    \centering
    \begin{tabular}{p{7cm}}
    \toprule
    You are a content moderation system. Your task is to detect hate speech in a social media messages. \\
    \bottomrule
    \end{tabular}
    \caption{System prompt.}
    \label{tab:system-prompt}
\end{table}
    
\begin{table}[ht!]
    \tabcolsep=0.08cm
    \centering
    \begin{tabular}{p{7cm}}
    \toprule
    Respond only with 'True' if the message contains hate speech, or 'False' if it does not. Do not explain, justify, or add anything else. Respond with exactly one word: True or False. \\
    \bottomrule
    \end{tabular}
    \caption{Base instruction.}
    \label{tab:base-instr}
\end{table}

\paragraph{Inference setup.} During inference, all models were run under a unified configuration to ensure comparability. We set the maximum sequence length to \num{4096} tokens to accommodate longer social media posts without truncation. The generation was constrained to a single output token ($max\_tokens = 1 $), since models were required to produce only one-word answers (\texttt{True} or \texttt{False}), that, for our selected models, correspond to a single token. To guarantee deterministic outputs, we disabled sampling by fixing $do\_sample = False$, and used a very low $temperature$ (0.01) together with $top\_p = 0.1$ and $top\_k = 5$. Finally, to prevent invalid generations, we restricted the output space using a custom \texttt{LogitsProcessor} from \texttt{transformers}\footnote{\url{https://huggingface.co/docs/transformers/}} library that filtered all tokens except \texttt{True} and \texttt{False}. Each LLM was executed in 4-bit quantized inference mode using \texttt{unsloth}\footnote{\url{https://unsloth.ai/}} framework. The resulting predictions constitute the \emph{LLM judgements}\footnote{Our code is available at \url{https://github.com/palomapiot/hate-eval-agreement/}}.

\paragraph{LLM pseudo-raters.} As we did with the human annotations, we also combined individual model predictions. We derived a majority label, based on the majority vote across LLMs, and also recorded whether this majority label was unanimous or not (LLM consensus).

\subsection{Metrics}

\paragraph{Agreement metrics.} To assess annotation quality and reliability, we compute both traditional and subjectivity-aware IAA measures. As standard metrics, we include \textbf{Cohen's $\kappa$}~\cite{doi:10.1177/001316446002000104}, \textbf{Fleiss's $\kappa$} \cite{Shrout1979} and \textbf{Krippendorff's $\alpha$} \cite{Hayes2007}. 

Cohen's $\kappa$ measures the degree of pairwise agreement between two annotators while correcting for agreement expected by chance. It is widely used in annotation studies to quantify consistency between raters and to distinguish systematic alignment from random coincidence. $\kappa$ values typically range from \num{0} (no agreement beyond chance) to \num{1} (perfect agreement), with intermediate ranges (e.g., \num{0.2}–\num{0.4}) often interpreted as fair or moderate agreement. Since our datasets were annotated by three annotators, we compute Cohen's $\kappa$ for all annotator pairs and report the average, which also enables compatibility with the xRR framework~\cite{Wong2021} (see below), as xRR builds upon pairwise $\kappa$ to quantify inter-group agreement. To complement pairwise analyses, we additionally compute Fleiss' $\kappa$ which generalises Cohen's $\kappa$ to multiple raters, and Krippendorff's $\alpha$, a flexible measure that accommodates multiple rates, missing data and different measurement levels.

These metrics all correct for chance agreement, assuming that all disagreements carry equal weight and that it exists a single, objective ground truth. These assumptions are often problematic in socially grounded tasks such as hate speech detection, where annotator background, cultural framing, or interpretation of intent can legitimately diverge. To address these limitations, we also employ \textbf{xRR} (cross-Rater Reliability) framework~\cite{Wong2021}, which extends traditional reliability analysis by accounting for systematic patterns of disagreement. xRR models the extent to which one group of annotators can reproduce the distribution of labels from another, offering a more nuanced and interpretable estimate of replicability in subjective settings. We omit metrics like simple percent agreement or Scott's $\pi$, as they either ignore chance or assume annotator interchangeability, making them unsuitable for subjective, categorical annotations. 



\section{Experiment 1: Agreement Analysis}
\label{sec:agreement}

\begin{figure*}[h]
    \centering
    \includegraphics[scale=0.18]{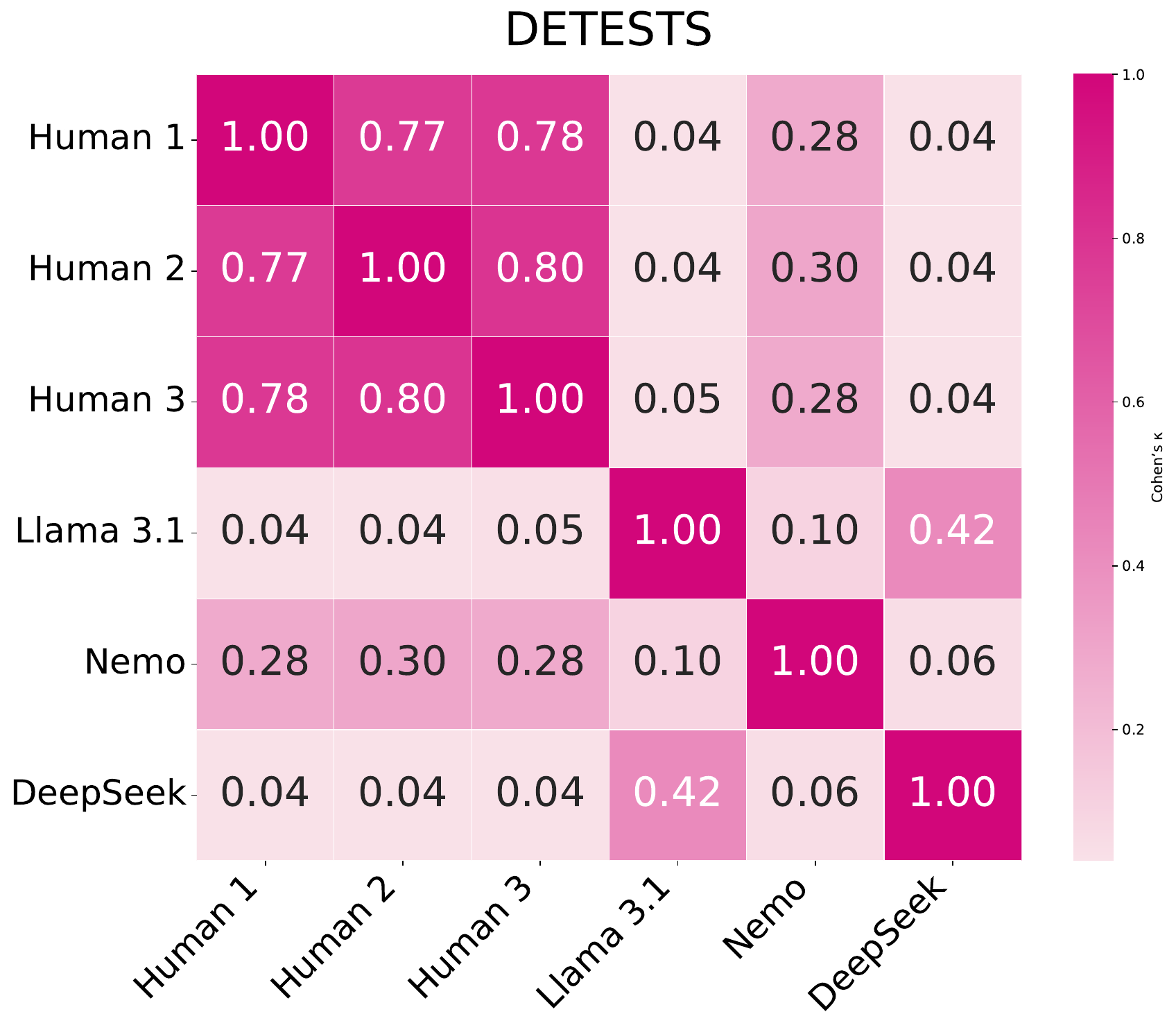}
    \includegraphics[scale=0.18]{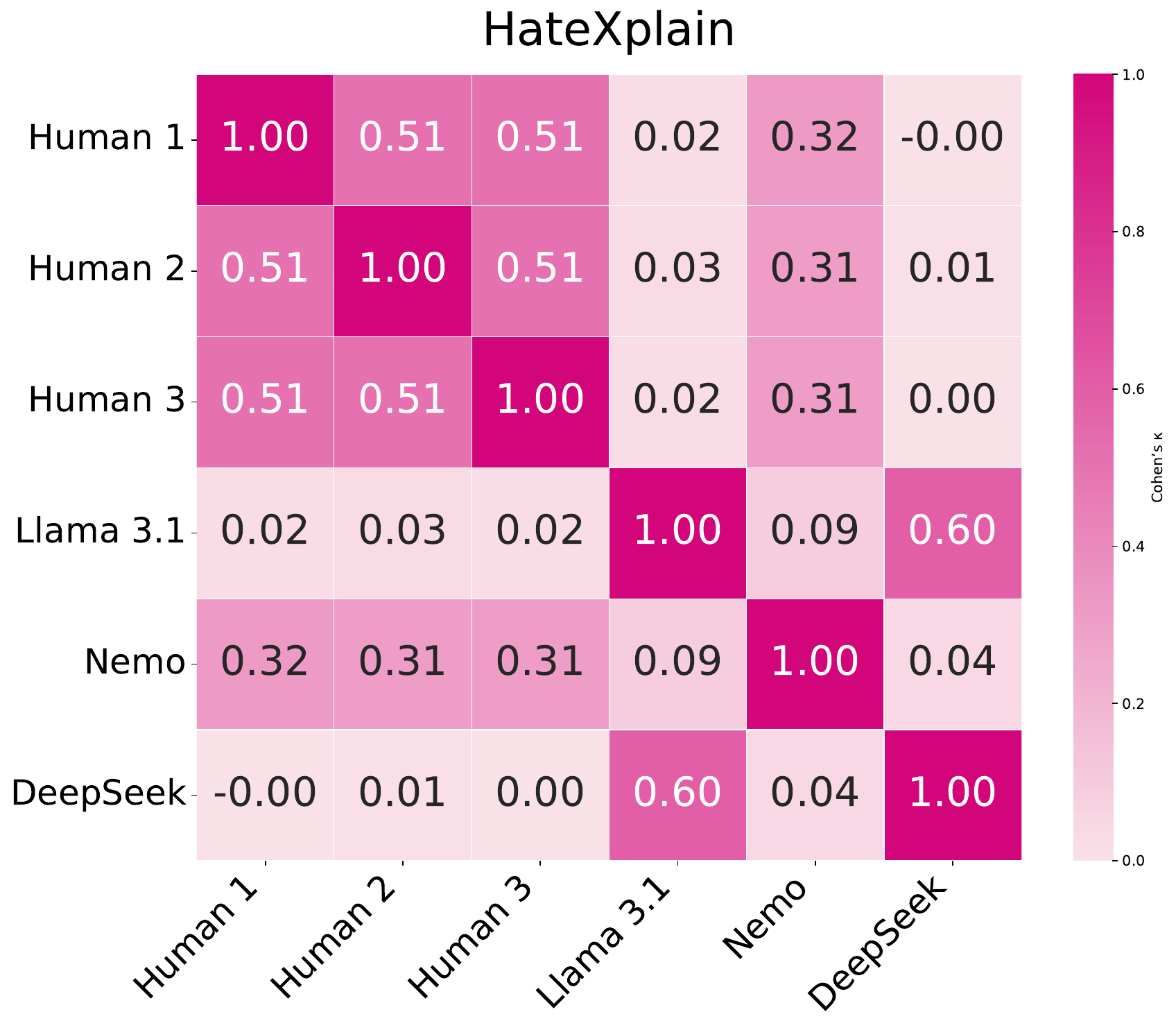}
    \includegraphics[scale=0.18]{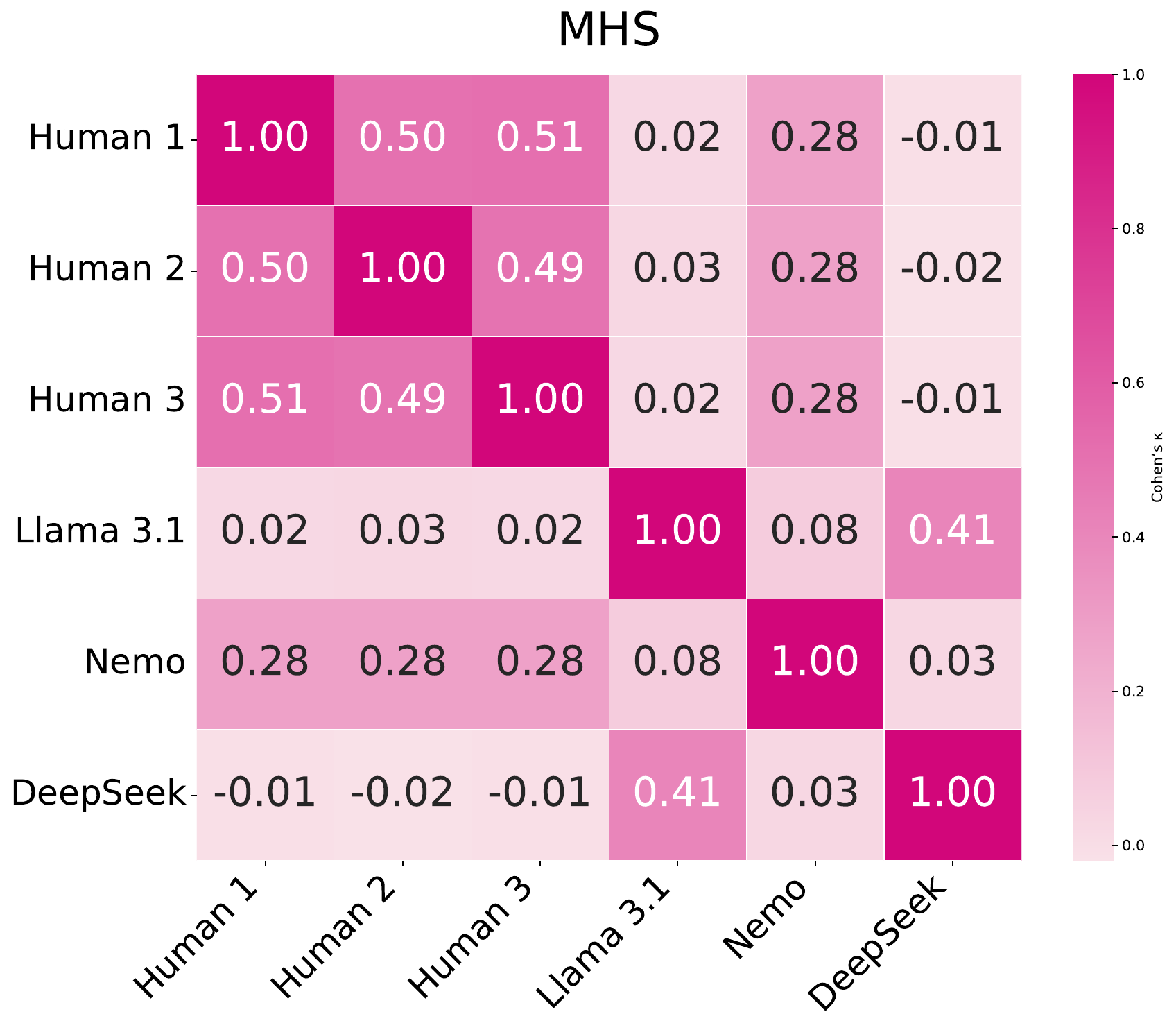}
    \caption{Pairwise Cohen's $\kappa$ agreement between human annotators and LLMs across the three datasets. Cells show $\kappa$ scores between pairs of raters.}
    \label{fig:kappa_heatmap}
\end{figure*}

The first experiment focuses on annotation agreement, addressing \textbf{RQ1} and \textbf{RQ2}. Our goal is to assess how well LLM-generated annotations align with human judgements, and whether subjectivity-aware metrics provide a more nuanced view than traditional measures.

\paragraph{Traditional metrics.}

We begin by assessing annotation reliability using Cohen's $\kappa$~\cite{doi:10.1177/001316446002000104}, a statistic that measures inter-rater reliability, assessing the agreement between two raters on categorical items while accounting for agreement that could occur by chance. To do so, we compared $(i)$ pairwise $\kappa$ between human annotators and LLMs, $(ii)$ each rater vs. the consensus of the other raters within the group (\textit{leave-one-out} $\kappa$), $(iii)$ each human vs. LLM majority and each LLM vs. human majority (cross-group $\kappa$), and $(iv)$ group-level $\kappa$ based on the majority and consensus labels. \Cref{fig:kappa_heatmap} reports $(i)$ and \Cref{tab:llm_cohen_kappa} reports the rest of the $\kappa$ scores. 


\begin{table}[ht]
    \centering
    \setlength{\tabcolsep}{6pt}
    \resizebox{\columnwidth}{!}{%
    \begin{tabular}{ll rrr}
        \toprule
        & \textbf{Rater} & \textbf{\textsc{DETESTS}} & \textbf{\textsc{HateXplain}} & \textbf{\textsc{MHS}} \\
        \midrule
        Leave-one-out: & \texttt{H1} & \num{0.840} & \num{0.470} & \num{0.487} \\
        \textit{Humans} & \texttt{H2} & \num{0.822} & \num{0.471} & \num{0.478} \\
        & \texttt{H3} & \num{0.852} & \num{0.462} & \num{0.478} \\
        \cmidrule(lr){2-5}
        & Mean & \num{0.838} & \num{0.468} & \num{0.481} \\
        \midrule
        Cross-group: & \texttt{Llama\phantom{0000}vs. H1+H2} & \num{0.052} & \num{0.027} & \num{0.019} \\
        \textit{LLM vs. humans} & \texttt{Llama\phantom{0000}vs. H1+H3} & \num{0.054} & \num{0.022} & \num{0.010} \\
        \textit{leave-one-out} & \texttt{Llama\phantom{0000}vs. H2+H3} & \num{0.051} & \num{0.027} & \num{0.025} \\
        & \texttt{Nemo\phantom{00000}vs. H1+H2} & $\boldsymbol{0.300}$ & $\boldsymbol{0.246}$ & $\boldsymbol{0.223}$ \\
        & \texttt{Nemo\phantom{00000}vs. H1+H3} & $\boldsymbol{0.293}$ & $\boldsymbol{0.243}$ & $\boldsymbol{0.223}$ \\
        & \texttt{Nemo\phantom{00000}vs. H2+H3} & $\boldsymbol{0.304}$ & $\boldsymbol{0.244}$ & $\boldsymbol{0.217}$ \\
        & \texttt{DeepSeek vs. H1+H2} & \num{0.043} & \num{0.001} & \num{-0.019} \\
        & \texttt{DeepSeek vs. H1+H3} & \num{0.045} & \num{-0.006} & \num{-0.018} \\
        & \texttt{DeepSeek vs. H2+H3} & \num{0.045} & \num{0.000} & \num{-0.018} \\
        \midrule
        Cross-group: & \texttt{Llama} & \num{0.048} & \num{0.029} & \num{0.023} \\
        \textit{LLM vs. humans maj.} & \texttt{Nemo} & $\boldsymbol{0.309}$ & $\boldsymbol{0.343}$ & $\boldsymbol{0.296}$ \\
        & \texttt{DeepSeek} & \num{0.042} & \num{0.005} & \num{-0.017} \\
        \cmidrule(lr){2-5}
        & Mean & \num{0.133} & \num{0.126} & \num{0.101} \\
        \midrule
        Cross-group:  & \texttt{H1} & \num{0.060} & \num{0.025} & \num{0.015} \\
        \textit{Human vs. LLMs maj.} & \texttt{H2} & \num{0.059} & \num{0.031} & \num{0.023} \\
        & \texttt{H3} & \num{0.061} & \num{0.024} & \num{0.018} \\
        \cmidrule(lr){2-5}
        & Mean & \num{0.060} & \num{0.026} & \num{0.019} \\
        \midrule
        Group-level & Majority & \num{0.066} & \num{0.029} & \num{0.020} \\
        & Consensus & \num{0.036} & \num{0.001} & \num{0.006} \\
        \bottomrule
    \end{tabular}
    }
    \caption{
    Cohen's $\kappa$ agreement scores across human annotators and LLMs for all datasets.  The table reports: $(ii)$ leave-one-out $\kappa$ (each rater vs. the consensus of others within their group), $(iii)$ cross-group $\kappa$ (each human vs. LLM majority and each LLM vs. human majority), and $(iv)$ group-level $\kappa$ based on majority and consensus labels. \textit{H} represents ``human''.}
    \label{tab:llm_cohen_kappa}
\end{table}

Human annotators exhibit high pairwise agreement across datasets, with mean inter-human $\kappa$ values ranging from \num{0.50} to \num{0.78}. The \textit{leave-one-out} analysis further confirms this reliability: each human annotator aligns closely with the consensus of the others, yielding $\kappa$ values between \num{0.46} and \num{0.85}. Among datasets, \textsc{DETESTS} achieves an almost perfect agreement ($\kappa\approx0.84$), while \textsc{HateXplain} and \textsc{MHS} remain at moderate agreement ($\kappa\approx0.47$). These results suggest that human annotators are relatively stable, even when evaluated ``\textit{one versus the rest}'' scenario, highlighting a robust signal in the underlying judgement process. 

In contrast, LLMs exhibit lower agreement both among themselves and with human annotators. Overall, pairwise inter-LLM $\kappa$ is generally \emph{slight} ($\kappa\approx0.17-0.24$ on average). When comparing LLMs with both the human majority and the \textit{leave-one-out} human evaluations a clear pattern emerges: \texttt{Nemo} demonstrates the most consistent alignment with humans, achieving $\kappa$ values near the \emph{fair} agreement range (\num{0.22}-\num{0.34}) in pairwise comparisons. These scores are comparable to those reported for crowdworkers or lightly instructed annotators in prior subjective annotation studies~\cite{hasoc19}. This suggests that \texttt{Nemo} is able to capture meaningful aspects of human annotations patterns, despite not being fine-tuned for the task. By contrast, \texttt{Llama 3.1} and \texttt{DeepSeek} show negligible agreement with the human majority and \textit{leave-one-out} human annotations (below \num{0.06}), with \texttt{DeepSeek} even exhibiting negative values. This suggests that these models are unreliable for hate speech annotation.  

Cross-group comparisons illustrate the persistent gap between human and LLM annotations. When each human is compared to the LLM majority, $\kappa$ values remain below \num{0.06} on average, whereas individual LLMs compared to the human majority achieve slightly higher agreement (up to \num{0.34} for \texttt{Nemo}).  Among the models, \texttt{Nemo} consistently shows the strongest alignment with human annotations, further confirming its comparatively higher sensitivity to human labelling patterns. These findings suggest that while aggregated LLM predictions (majority or consensus) slightly approach human alignment, they still fall far short of human reliability. The asymmetry between ``human vs. LLM majority'' and ``LLM vs. human majority'' underscores the uneven way LLMs capture human labelling patterns.

At the group level, $\kappa$ scores based on majority or full consensus are minimal ($<0.07$), confirming that aggregated pseudo-raters do not reach the reliability of human annotators. Overall, our analysis demonstrates that while LLMs can approximate certain trends in human annotations, the gap in agreement remains substantial, emphasising the need for careful interpretation of LLM-generated labels in subjective, socially grounded tasks such as hate speech detection.


\paragraph{Subjectivity-aware metrics.} To better capture the reliability of annotations in subjective settings like hate speech detection, we complement traditional agreement measures with xRR. This framework begins with a general definition of Cohen's $\kappa$ that is extended to cross-kappa ($\kappa_x$)--designed to measure annotation agreement between replications in a chance-corrected manner--and then is used to define $normalized~\kappa_x$, that measures \emph{similarity} between two replications~\cite{Wong2021}.  One of the utilities of this metric is to assess whether collected crowdsourced data closely mirror the target (i.e. data with expert annotations). In our study, the crowdsourced data are our LLMs judgements, and our target is the human-annotated data. According to the authors of the metric, a high $normalized~\kappa_x$ can assure us that the crowdsourced annotators (i.e. LLMs) are functioning as an extension of the trusted annotators. This metric approximates the true correlation between two experiments' item-level mean scores.

Table~\ref{tab:normalized_xrr} reports Fleiss' $\kappa$, Krippendorff's $\alpha$ and the average pairwise Cohen's $\kappa$ for both human and LLM annotators across datasets, alongside the $normalized~\kappa_x$ between Cohen's pairs. As expected, human annotators display strong internal agreement, with $\kappa,\alpha = 0.78$ for \textsc{DETESTS} and moderate values around $0.5$ for \textsc{HateXplain} and \textsc{MHS}. In contrast, the LLMs exhibits slight to fair agreement ($\kappa \approx 0.17-0.24$), indicating substantial divergence from humans.

Although LLMs exhibit only slight-to-fair agreement, the $normalized~\kappa_x$ values reveal a fair to moderate degree of similarity between the human and LLMs annotations ($0.35 \leq normalized~\kappa_x \leq 0.41$). This metric does not evaluate pairwise agreement within a group, but instead compares two independent replications of an annotation experiment (in our case, the human and the LLM groups). It quantifies how similarly the two groups behave on average across items, correcting for chance and for differences in the marginal label distributions. Hence, a moderate $normalized~\kappa_x$ indicates that, although the LLMs disagree among themselves on individual items, the overall pattern of their aggregated annotations moderately mirrors the general tendencies observed in the human judgements.  While traditional IAA metrics portray LLMs as largely unreliable, subjectivity-aware $normalized~\kappa_x$ shows a more positive picture: although LLMs still fall short of human-level reliability, they partially capture human annotation patterns.

\begin{table}
    \centering
    \resizebox{\columnwidth}{!}{%
    \begin{tabular}{lrrr}
        \toprule
         & \textbf{\textsc{DETESTS}} & \textbf{\textsc{HateXplain}} & \textbf{\textsc{MHS}} \\
        \midrule
        Human Fleiss's $\kappa$ & \num{0.784} & \num{0.510} & \num{0.501} \\
        LLMs Fleiss's $\kappa$ & \num{0.045} & \num{-0.032} & \num{-0.010} \\
        \midrule
        Human Krippendorffs's $\alpha$ & \num{0.784} & \num{0.510} & \num{0.501} \\ 
        LLMs Krippendorff's $\alpha$ & \num{0.045} & \num{-0.032} & \num{-0.010} \\
        \midrule
        Human mean Cohen's $\kappa$ & \num{0.784} & \num{0.510} & \num{0.501} \\
        LLMs mean Cohen's $\kappa$ & \num{0.195} & \num{0.244} & \num{0.174} \\
        \midrule
        $normalized~\kappa_x$ & \num{0.357} & \num{0.412} & \num{0.411} \\
    \bottomrule
    \end{tabular}
    }
    \caption{Inter-rater reliability for human annotators and LLMs, computed as using Fleiss' $\kappa$, Krippendorff's $\alpha$ and mean pairwise Cohen's $\kappa$, with $normalized~\kappa_x$ indicating the similarity between LLM and human annotation distributions. Metric values are reported up to three decimal; while they appear identical for humans, differences become apparent at the fifth or sixth decimal.}
    \label{tab:normalized_xrr}
\end{table}


\paragraph{Error analysis.} 

We conducted a qualitative error analysis on the \textsc{HateXplain} dataset, which was selected because it provides fine-grained, target-level annotations. This granularity allows us to identify which specific groups are systematically missed by LLMs. To capture overall trends, we computed per-target error rates, comparing both individual model predictions and the majority-vote aggregation against the human majority labels. 

First, we examine error distributions, where LLMs show a clear asymmetry. Both, \texttt{Llama 3.1} and \texttt{DeepSeek} miss a large number of hate instances (\num{2794} and \num{2896} false negatives, respectively) while generating comparatively few false positives (\num{691} and \num{447}), indicating a conservative detection bias. In contrast, \texttt{Nemo} attains the lowest false negative rate (\num{620}) but substantially over-predicts hate, with \num{7175} false positives. When considering the majority vote of all LLMs, the combined predictions yield \num{2819} false negatives and \num{594} false positives.

\begin{figure}[h]
    \centering
    \includegraphics[scale=0.33]{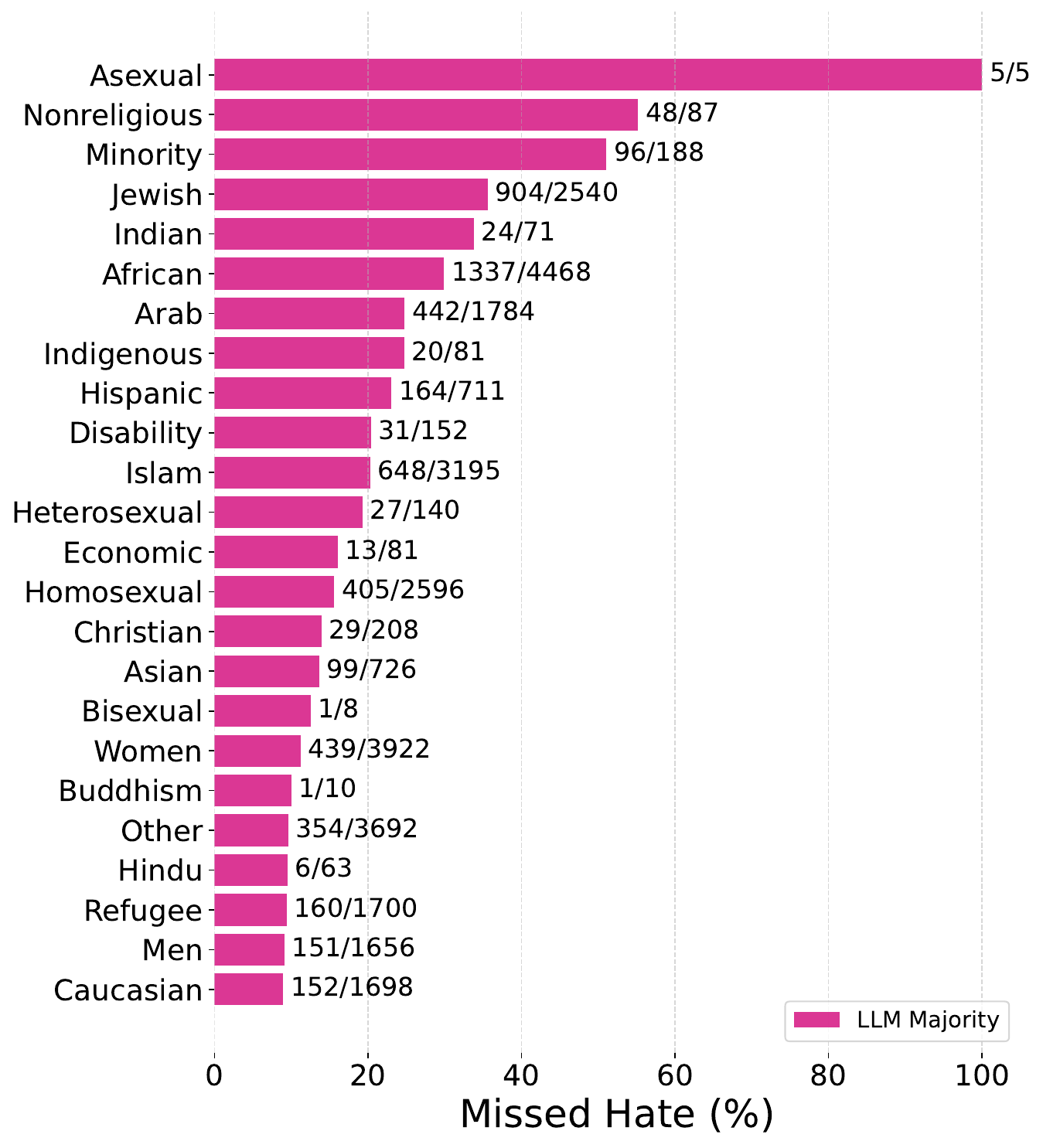}
    \caption{Per-target Missed Hate (\%) for LLM majority.}
    \label{fig:targets}
\end{figure}

\Cref{fig:targets} shows the the per-target analysis, highlighting systematic weaknesses of LLMs. Rare or nuanced targets such as \textit{Asexual} ($100\%$ missed), \textit{Non Religious} ($55.2\%$), \textit{Minority} ($51.1\%$) and \textit{Jewish} ($35.6\%$) are particularly challenging, reflecting under-detection for low-frequency groups (e.g. target \textit{Asexual} only has \num{5} posts in the dataset) or contextually subtle hate. Similarly, racial and ethnic categories like \textit{African} ($29.9\%$) and \textit{Arab} ($24.8\%$) also exhibit substantial false negatives. In contrast, more frequent or explicitly signalled targets such as \textit{Women} ($11.2\%$), \textit{Men} ($9.1\%$), \textit{Refugee} ($9.4\%$) and \textit{Caucasian} ($9.0\%$) show lower miss rates, indicating stronger detection of gender and migration-related hate. Religious categories such as \textit{Islam} ($20.3\%$) and \textit{Christian} ($13.9\%$) are captured moderately, suggesting partial coverage of faith-based hostility.

Across models, \texttt{Llama 3.1} and \texttt{DeepSeek} display near identical patterns, while \texttt{Nemo} shows lower miss rates across all targets, but often at the cost of over-predicting hate, as reflected in its high false positive count. Overall, these findings indicate that current LLMs effectively capture sexism and gender-related hate speech, but remain limited in recognising hate speech directed to other minority groups.

\section{Experiment 2: Ranking Correlation}


\Cref{sec:agreement} showed that LLMs are not yet reliable as annotators at the instance level. However, their partial alignment with human labelling behaviour motivates exploring an alternative role: their potential as system evaluators. In this second experiment we focus on \textbf{RQ3}, which asks whether LLM-generated labels can be leveraged to assess hate speech detection classifiers by preserving the relative ordering of model performance observed under human judgements. Unlike instance-level agreement, the primary concern for evaluation is whether the labels induce stable, human-consistent rankings across classifiers. For example, if we have three different classifiers, and evaluation under human annotations yields the ordering
$\texttt{Model A} > \texttt{Model B} > \texttt{Model C}$, and evaluation under LLM-based annotations yields the same ordering, then we would conclude that LLM-based annotations can be considered valid for evaluation.

\begin{table*}
    \centering
    \begin{tabular}{l rrrrrr}
        \toprule
        & \multicolumn{2}{c}{\textbf{\textsc{DETESTS}}} & \multicolumn{2}{c}{\textbf{\textsc{HateXplain}}} & \multicolumn{2}{c}{\textbf{\textsc{MHS}}} \\
        \cmidrule(lr){2-3} \cmidrule(lr){4-5} \cmidrule(lr){6-7}
        \textbf{Annotator} & $\tau$ & F1 Diff. & $\tau$ & F1 Diff. & $\tau$ & F1 Diff. \\
        \midrule
        LLM consensus & \num{0.898} & \num{0.381} & \num{0.757} & \num{0.386} & \num{0.617} & \num{0.376} \\
        LLM majority & \num{0.910} & \num{0.353} & \num{0.825} & \num{0.348} & \num{0.743} & \num{0.347} \\
        \texttt{Llama 3.1} & \num{0.931} & \num{0.357} & \num{0.846} & \num{0.338} & \num{0.788} & \num{0.351} \\
        \texttt{Nemo} & $\boldsymbol{0.952}$ & $\boldsymbol{0.041}$ & $\boldsymbol{0.964}$ & $\boldsymbol{-0.074}$ & $\boldsymbol{0.949}$ & $\boldsymbol{-0.006}$ \\
        \texttt{DeepSeek} & \num{0.893} & \num{0.342} & \num{0.419} & \num{0.370} & \num{-0.747} & \num{0.351} \\
        \midrule
        Human 1 & \num{0.997} & \num{-0.005} & \num{0.998} & \num{0.011} & \num{0.995} & \num{0.005} \\
        Human 2 & \num{0.997} & \num{-0.020} & \num{0.999} & \num{0.012} & \num{0.995} & \num{0.004} \\
        Human 3 & \num{0.998} & \num{-0.009} & \num{0.999} & \num{0.006} & \num{0.997} & \num{0.003} \\
        \bottomrule
    \end{tabular}
    \caption{Ranking correlation (Kendall's $\tau$) and mean F1 difference for human vs. LLM annotation. For each dataset, results for the best LLM are \textbf{in bold}.}
    \label{tab:correlation}
\end{table*}

\paragraph{Experimental setup.} To test this, we simulate a set of classifiers using a controlled degradation procedure. We start from an ``oracle system'' that perfectly reproduces the human majority labels ($F1 = 1.0$), and generate synthetic classifiers by randomly flipping a proportion $p\%$ of predictions. Each synthetic classifier is then evaluated against both human and LLM labels, producing paired performance scores. From these scores, we derive rankings of classifiers under human and LLM evaluation, which we compare using Kendall's $\tau$~\cite{Kendall1938}. Kendall's $\tau$ is a rank correlation coefficient that measures the similarity between two orderings: it ranges from $-1$ (inverse ordering) to $1$ (identical order), with $0$ indicating no association. To complement this ranking-based measure, we also report the mean absolute difference in F1\textsubscript{BINARY} scores across models to quantify potential distortions in absolute performance. \Cref{alg:correlation} summarises our experiment. With this setup we can assess whether LLM-generated annotations can reproduce human-derived performance orderings, even when per-instance agreement is limited, providing a practical test of their utility as evaluators.

\begin{algorithm}
\KwIn{Human labels $H$, LLM labels $L$, degradation levels $P = \{p_1, \dots, p_k\}$}
\KwOut{Kendall's $\tau$ correlation between rankings, AverageF1Diff}

\ForEach{$p \in P$}{
  Generate synthetic predictions $D_p$ by flipping $p\%$ of $H$ (gold labels $H$ themselves remain unchanged)\; 
  Compute $F1_H(p) \gets F1(H, D_p)$\; 
  Compute $F1_L(p) \gets F1(L, D_p)$\; 
  Store $(p, F1_H(p), F1_L(p))$ in results\;
}
Build rankings $Rank_H$ and $Rank_L$ from results \;
Compute Kendall's $\tau$ between $Rank_H$ and $Rank_L$\;
Compute $AverageF1Diff \gets \frac{1}{|P|} \sum_{p \in P} F1_H(p) - F1_L(p)$\;
\Return $(\tau, AverageF1Diff)$\;
\caption{Correlation-based Evaluation of LLM Annotations}
\label{alg:correlation}
\end{algorithm}

\paragraph{Results.} 
\Cref{tab:correlation} presents Kendall's $\tau$ between rankings derived from human and LLM labels, alongside the mean absolute difference in F1\textsubscript{BINARY} scores. Across datasets, most LLM variants achieve high rank correlation ($\tau \approx 0.84$-$0.96$), indicating that despite low instance-level agreement (see \Cref{tab:llm_cohen_kappa}), they largely preserve the relative order of classifiers. The strongest performer is \texttt{Nemo}, consistently achieving $\tau \ge 0.95$ with minimal F1\textsubscript{BINARY} deviation from human-based evaluation (e.g. $0.041$ on \textsc{DETESTS}). In contrast, \texttt{DeepSeek} fails on \textsc{MHS}, ($\tau < 0$), producing an inverted ranking relative to human judgements. For \texttt{Llama 3.1} and \texttt{DeepSeek}, mean F1\textsubscript{BINARY} differences remain substantial, showing that high rank correlation does not ensure accurate absolute scores. Overall, these results indicate that preserving relative model rankings is a less stringent requirement unlike instance agreement and can be achieved by some LLMs even when $\kappa$ is low. While LLMs offer coarse evaluations of relative model performance, their reliability is model and dataset dependent, and absolute score estimates remain biased.





\section{Conclusions}

This study examined the reliability of LLMs as annotators for hate speech detection through the lens of subjectivity-aware metrics. We compared traditional agreement metrics, such as Cohen's $\kappa$, with the cross-Rater Reliability framework, which accounts for systematic disagreement patterns across annotator groups. 

Our findings show that while LLMs achieve only slight to fair agreement under conventional metrics, subjectivity-aware evaluation paints a more optimistic picture: their judgements partially mirror human annotation tendencies, though they remain below human-level reliability. These results suggest that LLMs should not yet be treated as full substitutes for human annotators in socially grounded tasks. However, when repurposed as evaluators rather than annotators, LLMs can reliable reproduce relative model performance. In this correlation experiment, stronger models achieve near-human ordering (Kendall's $\tau \geq 0.95$) with minimal mean F1 distortion, whereas weaker or misaligned models may produce inverted orderings (negative $\tau$). This suggests that preserving relative model rankings is a weaker requirement than annotator-level agreement: carefully selected LLM label can support model comparison and evaluation audits when human labels are scarce, provided that correlation is verified on the target dataset.

Overall, LLMs are not yet reliable substitutes for human annotators in socially grounded tasks. Nonetheless, when treated as evaluators rather than annotators, they offer practical value: our correlation experiments show that LLMs can reproduce human-like model rankings with high consistency, even when their absolute judgements differ.

\section{Ethics}

Hate speech detection involves the processing of sensitive social media data and potentially harmful language. All datasets used in this study were publicly released and have been previously employed for research purposes. We strictly adhered to the terms of use of each dataset and ensured that no personally identifiable information was used or shared. While our experiments involved LLMs generating annotations oh hate-related content, we took care to avoid exposing harmful text outside controlled experimental settings. We acknowledge the societal risks of misclassification and the potential biases embedded in both human and machine judgements. Our work aims to promote transparency and responsible evaluation for sensitive tasks like hate speech detection.

\section{Limitations}

Our analysis is limited to binary hate speech detection and to datasets in English and Spanish, which may restrict the generalisability of our findings to other languages or more fine-grained hate categories. The LLMs evaluated represent only a subset of available architectures, and performance may vary with different models our prompt formulations. We intentionally adopted simple prompts to isolate model behaviour, leaving prompt variation outside the scope of this work. Finally, we constrained comparisons to instances with exactly three human annotators to ensure parity with the number of LLM raters, which may have reduced overall data coverage.

\section*{References}
\label{sec:reference}
\bibliographystyle{lrec2026-natbib}
\bibliography{references}

@inproceedings{Wong2021,
    title = "Cross-replication Reliability - An Empirical Approach to Interpreting Inter-rater Reliability",
    author = "Wong, Ka  and Paritosh, Praveen  and Aroyo, Lora",
    booktitle = "Proceedings of the 59th Annual Meeting of the Association for Computational Linguistics and the 11th International Joint Conference on Natural Language Processing (Volume 1: Long Papers)",
    year = "2021",
    publisher = "Association for Computational Linguistics",
    doi = "10.18653/v1/2021.acl-long.548",
    pages = "7053--7065",
}

@article{Piot2024,
  title={MetaHate: A Dataset for Unifying Efforts on Hate Speech Detection},
  volume={18},
  url={https://ojs.aaai.org/index.php/ICWSM/article/view/31445},
  DOI={10.1609/icwsm.v18i1.31445},
  number={1},
  journal={Proceedings of the International AAAI Conference on Web and Social Media},
  author={Piot, Paloma and Martín-Rodilla, Patricia and Parapar, Javier},
  year={2024},
  month={May},
  pages={2025-2039}
}

@article{GonzlezBailn2022,
  title = {Do social media undermine social cohesion? A critical review},
  volume = {17},
  ISSN = {1751-2409},
  url = {http://dx.doi.org/10.1111/sipr.12091},
  DOI = {10.1111/sipr.12091},
  number = {1},
  journal = {Social Issues and Policy Review},
  publisher = {Wiley},
  author = {González‐Bailón,  Sandra and Lelkes,  Yphtach},
  year = {2022},
  month = dec,
  pages = {155–180}
}

@article{davidsonhatespeech,
  title = {Automated Hate Speech Detection and the Problem of Offensive Language},
  volume = {11},
  ISSN = {2162-3449},
  DOI = {10.1609/icwsm.v11i1.14955},
  number = {1},
  journal = {Proceedings of the ICWSM 2017},
  publisher = {Association for the Advancement of Artificial Intelligence (AAAI)},
  author = {Davidson,  Thomas and Warmsley,  Dana and Macy,  Michael and Weber,  Ingmar},
  year = {2017},
  month = may,
  pages = {512–515}
}

@article{fountahatespeech,
  title = {Large Scale Crowdsourcing and Characterization of Twitter Abusive Behavior},
  volume = {12},
  ISSN = {2162-3449},
  DOI = {10.1609/icwsm.v12i1.14991},
  number = {1},
  journal = {Proceedings of the ICWSM 2018},
  publisher = {Association for the Advancement of Artificial Intelligence (AAAI)},
  author = {
    Founta,  Antigoni and 
    Djouvas,  Constantinos and
    Chatzakou,  Despoina and
    Leontiadis,  Ilias and
    Blackburn,  Jeremy and
    Stringhini,  Gianluca and
    ... and
    Kourtellis,  Nicolas},
  year = {2018},
  month = jun 
}

@inproceedings{mathewb2020hatexplain,
  title={HateXplain: A Benchmark Dataset for Explainable Hate Speech Detection},
  author={Binny Mathew and Punyajoy Saha and Seid Muhie Yimam and Chris Biemann and Pawan Goyal and Animesh Mukherjee},
  booktitle={Proceedings of the AAAI 2020},
  year={2020},
  url={https://api.semanticscholar.org/CorpusID:229332119}
}

@article{hatelingo2018elsherief,
  title = {Hate Lingo: A Target-Based Linguistic Analysis of Hate Speech in Social Media},
  volume = {12},
  ISSN = {2162-3449},
  DOI = {10.1609/icwsm.v12i1.15041},
  number = {1},
  journal = {Proceedings of the ICWSM 2018},
  publisher = {Association for the Advancement of Artificial Intelligence (AAAI)},
  author = {ElSherief,  Mai and Kulkarni,  Vivek and Nguyen,  Dana and Yang Wang,  William and Belding,  Elizabeth},
  year = {2018},
  month = jun 
}

@article{ElSherief2018,
  title = {Peer to Peer Hate: Hate Speech Instigators and Their Targets},
  volume = {12},
  ISSN = {2162-3449},
  DOI = {10.1609/icwsm.v12i1.15038},
  number = {1},
  journal = {Proceedings of the ICWSM 2018},
  publisher = {Association for the Advancement of Artificial Intelligence (AAAI)},
  author = {ElSherief,  Mai and Nilizadeh,  Shirin and Nguyen,  Dana and Vigna,  Giovanni and Belding,  Elizabeth},
  year = {2018},
  month = jun 
}

@article{hatemm2023,
  title = {HateMM: A Multi-Modal Dataset for Hate Video Classification},
  volume = {17},
  ISSN = {2162-3449},
  DOI = {10.1609/icwsm.v17i1.22209},
  journal = {Proceedings of the ICWSM 2023},
  publisher = {Association for the Advancement of Artificial Intelligence (AAAI)},
  author = {Das,  Mithun and Raj,  Rohit and Saha,  Punyajoy and Mathew,  Binny and Gupta,  Manish and Mukherjee,  Animesh},
  year = {2023},
  month = jun,
  pages = {1014–1023}
}

@inproceedings{glavas-etal-2020-xhate,
    title = "{XH}ate-999: Analyzing and Detecting Abusive Language Across Domains and Languages",
    author = "Glava{\v{s}}, Goran  and
      Karan, Vanja Mladen  and
      Vuli{\'c}, Ivan",
    editor = "Scott, Donia  and
      Bel, Nuria  and
      Zong, Chengqing",
    booktitle = "Proceedings of the 28th International Conference on Computational Linguistics",
    month = dec,
    year = "2020",
    address = "Barcelona, Spain (Online)",
    publisher = "International Committee on Computational Linguistics",
    url = "https://aclanthology.org/2020.coling-main.559/",
    doi = "10.18653/v1/2020.coling-main.559",
    pages = "6350--6365"
}

@inproceedings{plaza-del-arco-etal-2023-respectful,
    title = "Respectful or Toxic? Using Zero-Shot Learning with Language Models to Detect Hate Speech",
    author = "Plaza-del-arco, Flor Miriam  and
      Nozza, Debora  and
      Hovy, Dirk",
    editor = "Chung, Yi-ling  and
      R{\{}{\textbackslash}{''}ottger{\}}, Paul  and
      Nozza, Debora  and
      Talat, Zeerak  and
      Mostafazadeh Davani, Aida",
    booktitle = "The 7th Workshop on Online Abuse and Harms (WOAH)",
    month = jul,
    year = "2023",
    address = "Toronto, Canada",
    publisher = "Association for Computational Linguistics",
    url = "https://aclanthology.org/2023.woah-1.6/",
    doi = "10.18653/v1/2023.woah-1.6",
    pages = "60--68"
}

@inproceedings{roy-etal-2023-probing,
    title = "Probing {LLM}s for hate speech detection: strengths and vulnerabilities",
    author = "Roy, Sarthak  and
      Harshvardhan, Ashish  and
      Mukherjee, Animesh  and
      Saha, Punyajoy",
    editor = "Bouamor, Houda  and
      Pino, Juan  and
      Bali, Kalika",
    booktitle = "Findings of the Association for Computational Linguistics: EMNLP 2023",
    month = dec,
    year = "2023",
    address = "Singapore",
    publisher = "Association for Computational Linguistics",
    url = "https://aclanthology.org/2023.findings-emnlp.407/",
    doi = "10.18653/v1/2023.findings-emnlp.407",
    pages = "6116--6128"
}

@misc{wang2022toxicitydetectiongenerativepromptbased,
      title={Toxicity Detection with Generative Prompt-based Inference}, 
      author={Yau-Shian Wang and Yingshan Chang},
      year={2022},
      eprint={2205.12390},
      archivePrefix={arXiv},
      primaryClass={cs.CL},
      url={https://arxiv.org/abs/2205.12390}, 
}

@article{KansokDusche2022,
  title = {A Systematic Review on Hate Speech among Children and Adolescents: Definitions,  Prevalence,  and Overlap with Related Phenomena},
  volume = {24},
  ISSN = {1552-8324},
  url = {http://dx.doi.org/10.1177/15248380221108070},
  DOI = {10.1177/15248380221108070},
  number = {4},
  journal = {Trauma,  Violence,  \& Abuse},
  publisher = {SAGE Publications},
  author = {Kansok-Dusche,  Julia and Ballaschk,  Cindy and Krause,  Norman and Zeißig,  Anke and Seemann-Herz,  Lisanne and Wachs,  Sebastian and Bilz,  Ludwig},
  year = {2022},
  month = jun,
  pages = {2598–2615}
}

@misc{ADL2024,
  title = {Online Hate and Harassment: The American Experience 2024},
  author = {{Anti-Defamation League}},
  year = {2024},
  institution = {Anti-Defamation League},
  note = {Accessed: 03/01/2024},
  url = {https://www.adl.org/resources/report/online-hate-and-harassment-american-experience-2024}
}

@inbook{Piot2025,
  title = {{T}owards {E}fficient and {E}xplainable {H}ate {S}peech {D}etection via {M}odel {D}istillation},
  ISBN = {9783031887116},
  ISSN = {1611-3349},
  url = {http://dx.doi.org/10.1007/978-3-031-88711-6\_24},
  DOI = {10.1007/978-3-031-88711-6\_24},
  booktitle = {Advances in Information Retrieval},
  publisher = {Springer Nature Switzerland},
  author = {Piot,  Paloma and Parapar,  Javier},
  year = {2025},
  pages = {376–392}
}

@misc{grattafiori2024llama3herdmodels,
      title={The Llama 3 Herd of Models}, 
      author={Aaron Grattafiori and Abhimanyu Dubey and Abhinav Jauhri and Abhinav Pandey and Abhishek Kadian and Ahmad Al-Dahle and ... and Zhiyu Ma},
      year={2024},
      eprint={2407.21783},
      archivePrefix={arXiv},
      primaryClass={cs.AI},
      url={https://arxiv.org/abs/2407.21783}, 
}

@misc{mistralnemo,
	author = {{Mistral AI team}},
	title = {{M}istral {N}e{M}o | {M}istral {A}{I}},
	howpublished = {\url{https://mistral.ai/news/mistral-nemo}},
	year = {2025},
	note = {Accessed: 13/08/2025},
}

@article{Ljubei2022,
  title = {Quantifying the impact of context on the quality of manual hate speech annotation},
  volume = {29},
  ISSN = {1469-8110},
  url = {http://dx.doi.org/10.1017/S1351324922000353},
  DOI = {10.1017/s1351324922000353},
  number = {6},
  journal = {Natural Language Engineering},
  publisher = {Cambridge University Press (CUP)},
  author = {Ljubešić,  Nikola and Mozetič,  Igor and Kralj Novak,  Petra},
  year = {2022},
  month = aug,
  pages = {1481–1494}
}

@inproceedings{toraman-etal-2022-large,
    title = "Large-Scale Hate Speech Detection with Cross-Domain Transfer",
    author = "Toraman, Cagri  and
      {\c{S}}ahinu{\c{c}}, Furkan  and
      Yilmaz, Eyup",
    editor = "Calzolari, Nicoletta  and
      B{\'e}chet, Fr{\'e}d{\'e}ric  and
      Blache, Philippe  and
      Choukri, Khalid  and
      Cieri, Christopher  and
      Declerck, Thierry  and
      Goggi, Sara  and
      Isahara, Hitoshi  and
      Maegaard, Bente  and
      Mariani, Joseph  and
      Mazo, H{\'e}l{\`e}ne  and
      Odijk, Jan  and
      Piperidis, Stelios",
    booktitle = "Proceedings of the Thirteenth Language Resources and Evaluation Conference",
    month = jun,
    year = "2022",
    address = "Marseille, France",
    publisher = "European Language Resources Association",
    url = "https://aclanthology.org/2022.lrec-1.238/",
    pages = "2215--2225"
}

@inproceedings{barbarestani-etal-2024-content,
    title = "Content Moderation in Online Platforms: A Study of Annotation Methods for Inappropriate Language",
    author = "Barbarestani, Baran  and
      Maks, Isa  and
      Vossen, Piek T.J.M.",
    editor = "Kumar, Ritesh  and
      Ojha, Atul Kr.  and
      Malmasi, Shervin  and
      Chakravarthi, Bharathi Raja  and
      Lahiri, Bornini  and
      Singh, Siddharth  and
      Ratan, Shyam",
    booktitle = "Proceedings of the Fourth Workshop on Threat, Aggression {\&} Cyberbullying @ LREC-COLING-2024",
    month = may,
    year = "2024",
    address = "Torino, Italia",
    publisher = "ELRA and ICCL",
    url = "https://aclanthology.org/2024.trac-1.11/",
    pages = "96--104"
}

@inproceedings{kanclerz-etal-2022-ground,
    title = "What If Ground Truth Is Subjective? Personalized Deep Neural Hate Speech Detection",
    author = "Kanclerz, Kamil  and
      Gruza, Marcin  and
      Karanowski, Konrad  and
      Bielaniewicz, Julita  and
      Milkowski, Piotr  and
      Kocon, Jan  and
      Kazienko, Przemyslaw",
    editor = "Abercrombie, Gavin  and
      Basile, Valerio  and
      Tonelli, Sara  and
      Rieser, Verena  and
      Uma, Alexandra",
    booktitle = "Proceedings of the 1st Workshop on Perspectivist Approaches to NLP @LREC2022",
    month = jun,
    year = "2022",
    address = "Marseille, France",
    publisher = "European Language Resources Association",
    url = "https://aclanthology.org/2022.nlperspectives-1.6/",
    pages = "37--45"
}

@inproceedings{10.1145/3308560.3317083,
author = {Aroyo, Lora and Dixon, Lucas and Thain, Nithum and Redfield, Olivia and Rosen, Rachel},
title = {Crowdsourcing Subjective Tasks: The Case Study of Understanding Toxicity in Online Discussions},
year = {2019},
isbn = {9781450366755},
publisher = {Association for Computing Machinery},
address = {New York, NY, USA},
url = {https://doi.org/10.1145/3308560.3317083},
doi = {10.1145/3308560.3317083},
booktitle = {Companion Proceedings of The 2019 World Wide Web Conference},
pages = {1100–1105},
numpages = {6},
location = {San Francisco, USA},
series = {WWW '19}
}

@article{Hettiachchi2023, title={How Crowd Worker Factors Influence Subjective Annotations: A Study of Tagging Misogynistic Hate Speech in Tweets}, volume={11}, url={https://ojs.aaai.org/index.php/HCOMP/article/view/27546}, DOI={10.1609/hcomp.v11i1.27546}, abstractNote={Crowdsourced annotation is vital to both collecting labelled data to train and test automated content moderation systems and to support human-in-the-loop review of system decisions. However, annotation tasks such as judging hate speech are subjective and thus highly sensitive to biases stemming from annotator beliefs, characteristics and demographics. We conduct two crowdsourcing studies on Mechanical Turk to examine annotator bias in labelling sexist and misogynistic hate speech. Results from 109 annotators show that annotator political inclination, moral integrity, personality traits, and sexist attitudes significantly impact annotation accuracy and the tendency to tag content as hate speech. In addition, semi-structured interviews with nine crowd workers provide further insights regarding the influence of subjectivity on annotations. In exploring how workers interpret a task — shaped by complex negotiations between platform structures, task instructions, subjective motivations, and external contextual factors — we see annotations not only impacted by worker factors but also simultaneously shaped by the structures under which they labour.}, number={1}, journal={Proceedings of the AAAI Conference on Human Computation and Crowdsourcing}, author={Hettiachchi, Danula and Holcombe-James, Indigo and Livingstone, Stephanie and de Silva, Anjalee and Lease, Matthew and Salim, Flora D. and Sanderson, Mark}, year={2023}, month={Nov.}, pages={38-50} }

@misc{dehghan2025dealingannotatordisagreementhate,
      title={Dealing with Annotator Disagreement in Hate Speech Classification}, 
      author={Somaiyeh Dehghan and Mehmet Umut Sen and Berrin Yanikoglu},
      year={2025},
      eprint={2502.08266},
      archivePrefix={arXiv},
      primaryClass={cs.CL},
      url={https://arxiv.org/abs/2502.08266}, 
}

@inproceedings{rottger-etal-2022-two,
    title = "Two Contrasting Data Annotation Paradigms for Subjective {NLP} Tasks",
    author = {R{\"o}ttger, Paul  and
      Vidgen, Bertie  and
      Hovy, Dirk  and
      Pierrehumbert, Janet},
    editor = "Carpuat, Marine  and
      de Marneffe, Marie-Catherine  and
      Meza Ruiz, Ivan Vladimir",
    booktitle = "Proceedings of the 2022 Conference of the North American Chapter of the Association for Computational Linguistics: Human Language Technologies",
    month = jul,
    year = "2022",
    address = "Seattle, United States",
    publisher = "Association for Computational Linguistics",
    url = "https://aclanthology.org/2022.naacl-main.13/",
    doi = "10.18653/v1/2022.naacl-main.13",
    pages = "175--190"
}

@article{davani-etal-2023-hate,
    title = "Hate Speech Classifiers Learn Normative Social Stereotypes",
    author = "Davani, Aida Mostafazadeh  and
      Atari, Mohammad  and
      Kennedy, Brendan  and
      Dehghani, Morteza",
    journal = "Transactions of the Association for Computational Linguistics",
    volume = "11",
    year = "2023",
    address = "Cambridge, MA",
    publisher = "MIT Press",
    url = "https://aclanthology.org/2023.tacl-1.18/",
    doi = "10.1162/tacl_a_00550",
    pages = "300--319"
}

@InProceedings{10.1007/978-3-031-08974-9_54,
author="Kralj Novak, Petra
and Scantamburlo, Teresa
and Pelicon, Andra{\v{z}}
and Cinelli, Matteo
and Mozeti{\v{c}}, Igor
and Zollo, Fabiana",
editor="Ciucci, Davide
and Couso, In{\'e}s
and Medina, Jes{\'u}s
and {\'{S}}l{\k{e}}zak, Dominik
and Petturiti, Davide
and Bouchon-Meunier, Bernadette
and Yager, Ronald R.",
title="Handling Disagreement in Hate Speech Modelling",
booktitle="Information Processing and Management of Uncertainty in Knowledge-Based Systems",
year="2022",
publisher="Springer International Publishing",
address="Cham",
pages="681--695",
isbn="978-3-031-08974-9"
}

@article{davani-etal-2022-dealing,
    title = "Dealing with Disagreements: Looking Beyond the Majority Vote in Subjective Annotations",
    author = "Mostafazadeh Davani, Aida  and
      D{\'i}az, Mark  and
      Prabhakaran, Vinodkumar",
    editor = "Roark, Brian  and
      Nenkova, Ani",
    journal = "Transactions of the Association for Computational Linguistics",
    volume = "10",
    year = "2022",
    address = "Cambridge, MA",
    publisher = "MIT Press",
    url = "https://aclanthology.org/2022.tacl-1.6/",
    doi = "10.1162/tacl_a_00449",
    pages = "92--110"
}

@article{doi:10.1177/001316446002000104,
    author = {Jacob Cohen},
    title ={A Coefficient of Agreement for Nominal Scales},
    journal = {Educational and Psychological Measurement},
    volume = {20},
    number = {1},
    pages = {37-46},
    year = {1960},
    doi = {10.1177/001316446002000104},
    URL = { "https://doi.org/10.1177/001316446002000104"
    },
    eprint = { "https://doi.org/10.1177/001316446002000104"
    }
}

@inproceedings{marchal-etal-2022-establishing,
    title = "Establishing Annotation Quality in Multi-label Annotations",
    author = "Marchal, Marian  and
      Scholman, Merel  and
      Yung, Frances  and
      Demberg, Vera",
    editor = "Calzolari, Nicoletta  and
      Huang, Chu-Ren  and
      Kim, Hansaem  and
      Pustejovsky, James  and
      Wanner, Leo  and
      Choi, Key-Sun  and
      Ryu, Pum-Mo  and
      Chen, Hsin-Hsi  and
      Donatelli, Lucia  and
      Ji, Heng  and
      Kurohashi, Sadao  and
      Paggio, Patrizia  and
      Xue, Nianwen  and
      Kim, Seokhwan  and
      Hahm, Younggyun  and
      He, Zhong  and
      Lee, Tony Kyungil  and
      Santus, Enrico  and
      Bond, Francis  and
      Na, Seung-Hoon",
    booktitle = "Proceedings of the 29th International Conference on Computational Linguistics",
    month = oct,
    year = "2022",
    address = "Gyeongju, Republic of Korea",
    publisher = "International Committee on Computational Linguistics",
    url = "https://aclanthology.org/2022.coling-1.322/",
    pages = "3659--3668"
}

@inproceedings{plank-2022-problem,
    title = "The ``Problem'' of Human Label Variation: On Ground Truth in Data, Modeling and Evaluation",
    author = "Plank, Barbara",
    editor = "Goldberg, Yoav  and
      Kozareva, Zornitsa  and
      Zhang, Yue",
    booktitle = "Proceedings of the 2022 Conference on Empirical Methods in Natural Language Processing",
    month = dec,
    year = "2022",
    address = "Abu Dhabi, United Arab Emirates",
    publisher = "Association for Computational Linguistics",
    url = "https://aclanthology.org/2022.emnlp-main.731/",
    doi = "10.18653/v1/2022.emnlp-main.731",
    pages = "10671--10682"
}

@misc{baumann2025largelanguagemodelhacking,
      title={Large Language Model Hacking: Quantifying the Hidden Risks of Using LLMs for Text Annotation}, 
      author={Joachim Baumann and Paul Röttger and Aleksandra Urman and Albert Wendsjö and Flor Miriam Plaza-del-Arco and Johannes B. Gruber and Dirk Hovy},
      year={2025},
      eprint={2509.08825},
      archivePrefix={arXiv},
      primaryClass={cs.CL},
      url={https://arxiv.org/abs/2509.08825}, 
}

@article{doi:10.1073/pnas.2305016120,
author = {Fabrizio Gilardi  and Meysam Alizadeh  and Maël Kubli },
title = {ChatGPT outperforms crowd workers for text-annotation tasks},
journal = {Proceedings of the National Academy of Sciences},
volume = {120},
number = {30},
pages = {e2305016120},
year = {2023},
doi = {10.1073/pnas.2305016120},
URL = {https://www.pnas.org/doi/abs/10.1073/pnas.2305016120},
eprint = {https://www.pnas.org/doi/pdf/10.1073/pnas.2305016120}}

@article{doi:10.1177/08944393241286471,
author = {Petter Törnberg},
title ={Large Language Models Outperform Expert Coders and Supervised Classifiers at Annotating Political Social Media Messages},
journal = {Social Science Computer Review},
volume = {0},
number = {0},
pages = {08944393241286471},
year = {2024},
doi = {10.1177/08944393241286471},
URL = { 
        https://doi.org/10.1177/08944393241286471
},
eprint = { 
        https://doi.org/10.1177/08944393241286471
}
}

@misc{tseng2025evaluatinglargelanguagemodels,
      title={Evaluating Large Language Models as Expert Annotators}, 
      author={Yu-Min Tseng and Wei-Lin Chen and Chung-Chi Chen and Hsin-Hsi Chen},
      year={2025},
      eprint={2508.07827},
      archivePrefix={arXiv},
      primaryClass={cs.CL},
      url={https://arxiv.org/abs/2508.07827}, 
}

@InProceedings{pmlr-v239-mohta23a,
  title = 	 {Are large language models good annotators?},
  author =       {Mohta, Jay and Ak, Kenan and Xu, Yan and Shen, Mingwei},
  booktitle = 	 {Proceedings on "I Can't Believe It's Not Better: Failure  Modes in the Age of Foundation Models" at NeurIPS 2023 Workshops},
  pages = 	 {38--48},
  year = 	 {2023},
  editor = 	 {Antorán, Javier and Blaas, Arno and Buchanan, Kelly and Feng, Fan and Fortuin, Vincent and Ghalebikesabi, Sahra and Kriegler, Andreas and Mason, Ian and Rohde, David and Ruiz, Francisco J. R. and Uelwer, Tobias and Xie, Yubin and Yang, Rui},
  volume = 	 {239},
  series = 	 {Proceedings of Machine Learning Research},
  month = 	 {16 Dec},
  publisher =    {PMLR},
  url = 	 {https://proceedings.mlr.press/v239/mohta23a.html}
}

@article{Giorgi2025, title={Human and LLM Biases in Hate Speech Annotations: A Socio-Demographic Analysis of Annotators and Targets}, volume={19}, url={https://ojs.aaai.org/index.php/ICWSM/article/view/35837}, DOI={10.1609/icwsm.v19i1.35837}, abstractNote={The rise of online platforms exacerbated the spread of hate speech, demanding scalable and effective detection. However, the accuracy of hate speech detection systems heavily relies on human-labeled data, which is inherently susceptible to biases. While previous work has examined the issue, the interplay between the characteristics of the annotator and those of the target of the hate are still unexplored. We fill this gap by leveraging an extensive dataset with rich socio-demographic information of both annotators and targets, uncovering how human biases manifest in relation to the target’s attributes. Our analysis surfaces the presence of widespread biases, which we quantitatively describe and characterize based on their intensity and prevalence, revealing marked differences. Furthermore, we compare human biases with those exhibited by persona-based LLMs. Our findings indicate that while persona-based LLMs do exhibit biases, these differ significantly from those of human annotators. Overall, our work offers new and nuanced results on human biases in hate speech annotations, as well as fresh insights into the design of AI-driven hate speech detection systems.}, number={1}, journal={Proceedings of the International AAAI Conference on Web and Social Media}, author={Giorgi, Tommaso and Cima, Lorenzo and Fagni, Tiziano and Avvenuti, Marco and Cresci, Stefano}, year={2025}, month={Jun.}, pages={653-670} }

@article{Matter2024,
  title = {Investigating the increase of violent speech in Incel communities with human-guided GPT-4 prompt iteration},
  volume = {2},
  ISSN = {2813-7876},
  url = {http://dx.doi.org/10.3389/frsps.2024.1383152},
  DOI = {10.3389/frsps.2024.1383152},
  journal = {Frontiers in Social Psychology},
  publisher = {Frontiers Media SA},
  author = {Matter,  Daniel and Schirmer,  Miriam and Grinberg,  Nir and Pfeffer,  J\"{u}rgen},
  year = {2024},
  month = jul 
}

@inproceedings{bassi-etal-2025-annotating,
    title = "Annotating the Annotators: Analysis, Insights and Modelling from an Annotation Campaign on Persuasion Techniques Detection",
    author = "Bassi, Davide  and
      Dimitrov, Dimitar Iliyanov  and
      D{'}Auria, Bernardo  and
      Alam, Firoj  and
      Hasanain, Maram  and
      Moro, Christian  and
      Orr{\`u}, Luisa  and
      Turchi, Gian Piero  and
      Nakov, Preslav  and
      Da San Martino, Giovanni",
    editor = "Che, Wanxiang  and
      Nabende, Joyce  and
      Shutova, Ekaterina  and
      Pilehvar, Mohammad Taher",
    booktitle = "Findings of the Association for Computational Linguistics: ACL 2025",
    month = jul,
    year = "2025",
    address = "Vienna, Austria",
    publisher = "Association for Computational Linguistics",
    url = "https://aclanthology.org/2025.findings-acl.922/",
    doi = "10.18653/v1/2025.findings-acl.922",
    pages = "17918--17929",
    ISBN = "979-8-89176-256-5"
}

@misc{kennedy2020constructingintervalvariablesfaceted,
      title={Constructing interval variables via faceted Rasch measurement and multitask deep learning: a hate speech application}, 
      author={Chris J. Kennedy and Geoff Bacon and Alexander Sahn and Claudia von Vacano},
      year={2020},
      eprint={2009.10277},
      archivePrefix={arXiv},
      primaryClass={cs.CL},
      url={https://arxiv.org/abs/2009.10277}, 
}

@inproceedings{sachdeva-etal-2022-measuring,
    title = "The Measuring Hate Speech Corpus: Leveraging Rasch Measurement Theory for Data Perspectivism",
    author = "Sachdeva, Pratik  and
      Barreto, Renata  and
      Bacon, Geoff  and
      Sahn, Alexander  and
      von Vacano, Claudia  and
      Kennedy, Chris",
    editor = "Abercrombie, Gavin  and
      Basile, Valerio  and
      Tonelli, Sara  and
      Rieser, Verena  and
      Uma, Alexandra",
    booktitle = "Proceedings of the 1st Workshop on Perspectivist Approaches to NLP @LREC2022",
    month = jun,
    year = "2022",
    address = "Marseille, France",
    publisher = "European Language Resources Association",
    url = "https://aclanthology.org/2022.nlperspectives-1.11/",
    pages = "83--94"
}

@article{exist2022,
  address = {ES},
  title = {Overview of EXIST 2022: sEXism Identification in Social neTworks},
  ISSN = {1989-7553},
  url = {https://doi.org/10.26342/2022-69-20},
  DOI = {10.26342/2022-69-20},
  journal = {Procesamiento del Lenguaje Natural},
  publisher = {Sociedad Española para el Procesamiento del Lenguaje Natural},
  author = {Rodríguez-Sánchez,  Francisco and Carrillo-de-Albornoz,  Jorge and Plaza,  Laura and Mendieta-Aragón,  Adrián and Marco-Remón,  Guillermo and Makeienko,  Maryna and Plaza,  María and Gonzalo,  Julio and Spina,  Damiano and Rosso,  Paolo},
  year = {2022},
  pages = {229–240}
}

@article{detests2024,
  title = {Overview of DETESTS-Dis at IberLEF 2024:
DETEction and classification of racial STereotypes in
Spanish - Learning with Disagreement},
  volume = {73},
  ISSN = {1135-5948},
  DOI = {10.26342/2024-73-24},
  journal = {Procesamiento del Lenguaje Natural},
  publisher = {Sociedad Española para el Procesamiento de Lenguaje Natural},
  author = {Schmeisser-Nieto, Wolfgang S. and
  Pasterlls, Pol and
  Frenda, Simona and
  Ariza-Casabona, Alejandro and
  Farrús, Mireia and
  Rosso, Paolo and
  Tulé, Mariona},
  year = {2024},
  month = sep,
  pages = {323-333}
}

@misc{deepseekai2025deepseekr1incentivizingreasoningcapability,
      title={DeepSeek-R1: Incentivizing Reasoning Capability in LLMs via Reinforcement Learning}, 
      author={DeepSeek-AI},
      year={2025},
      eprint={2501.12948},
      archivePrefix={arXiv},
      primaryClass={cs.CL},
      url={https://arxiv.org/abs/2501.12948}, 
}

@article{hasoc19,
	author = {
	Modha, Sandip and 
	Mandl, Thomas and 
	Majumder, Prasenjit and 
	Patel, Daksh},
	title = {Overview of the HASOC track at FIRE 2019: Hate Speech and Offensive Content Identification in Indo-European Languages},
	volume = {2517},
	ISSN = {1613-0073},
	journal = {Working Notes of FIRE 2019},
	publisher = {CEUR Workshop Proceedings},
	year = {2019},
}

@article{Shrout1979,
  title = {Intraclass correlations: Uses in assessing rater reliability.},
  volume = {86},
  ISSN = {0033-2909},
  url = {http://dx.doi.org/10.1037/0033-2909.86.2.420},
  DOI = {10.1037/0033-2909.86.2.420},
  number = {2},
  journal = {Psychological Bulletin},
  publisher = {American Psychological Association (APA)},
  author = {Shrout,  Patrick E. and Fleiss,  Joseph L.},
  year = {1979},
  pages = {420–428}
}

@article{10.1145/3232676,
author = {Fortuna, Paula and Nunes, S\'{e}rgio},
title = {A Survey on Automatic Detection of Hate Speech in Text},
year = {2018},
issue_date = {July 2019},
publisher = {Association for Computing Machinery},
address = {New York, NY, USA},
volume = {51},
number = {4},
issn = {0360-0300},
url = {https://doi.org/10.1145/3232676},
doi = {10.1145/3232676},
journal = {ACM Comput. Surv.},
month = jul,
articleno = {85},
numpages = {30}
}

@inproceedings{10.1007/978-3-030-77091-4_26,
author = {Basile, Valerio},
title = {It’s the End of the Gold Standard as We Know It: Leveraging Non-aggregated Data for Better Evaluation and Explanation of Subjective Tasks},
year = {2020},
isbn = {978-3-030-77090-7},
publisher = {Springer-Verlag},
address = {Berlin, Heidelberg},
url = {https://doi.org/10.1007/978-3-030-77091-4_26},
doi = {10.1007/978-3-030-77091-4_26},
booktitle = {AIxIA 2020 – Advances in Artificial Intelligence: XIXth International Conference of the Italian Association for Artificial Intelligence, Virtual Event,  November 25–27, 2020, Revised Selected Papers},
pages = {441–453},
numpages = {13},
location = {Milan, Italy}
}

@inproceedings{guellil-etal-2024-annotators,
    title = "The Annotators Agree to Not Agree on the Fine-grained Annotation of Hate-speech against Women in {A}lgerian Dialect Comments",
    author = "Guellil, Imane  and
      Houichi, Yousra  and
      Chennoufi, Sara  and
      Boubred, Mohamed  and
      Boucetta, Anfal Yousra  and
      Azouaou, Faical",
    editor = "Mabuya, Rooweither  and
      Matfunjwa, Muzi  and
      Setaka, Mmasibidi  and
      van Zaanen, Menno",
    booktitle = "Proceedings of the Fifth Workshop on Resources for African Indigenous Languages @ LREC-COLING 2024",
    month = may,
    year = "2024",
    address = "Torino, Italia",
    publisher = "ELRA and ICCL",
    url = "https://aclanthology.org/2024.rail-1.15/",
    pages = "133--139"
}

@inproceedings{leonardelli-etal-2021-agreeing,
    title = "Agreeing to Disagree: Annotating Offensive Language Datasets with Annotators' Disagreement",
    author = "Leonardelli, Elisa  and
      Menini, Stefano  and
      Palmero Aprosio, Alessio  and
      Guerini, Marco  and
      Tonelli, Sara",
    editor = "Moens, Marie-Francine  and
      Huang, Xuanjing  and
      Specia, Lucia  and
      Yih, Scott Wen-tau",
    booktitle = "Proceedings of the 2021 Conference on Empirical Methods in Natural Language Processing",
    month = nov,
    year = "2021",
    address = "Online and Punta Cana, Dominican Republic",
    publisher = "Association for Computational Linguistics",
    url = "https://aclanthology.org/2021.emnlp-main.822/",
    doi = "10.18653/v1/2021.emnlp-main.822",
    pages = "10528--10539"
}

@article{pavlick-kwiatkowski-2019-inherent,
    title = "Inherent Disagreements in Human Textual Inferences",
    author = "Pavlick, Ellie  and
      Kwiatkowski, Tom",
    editor = "Lee, Lillian  and
      Johnson, Mark  and
      Roark, Brian  and
      Nenkova, Ani",
    journal = "Transactions of the Association for Computational Linguistics",
    volume = "7",
    year = "2019",
    address = "Cambridge, MA",
    publisher = "MIT Press",
    url = "https://aclanthology.org/Q19-1043/",
    doi = "10.1162/tacl_a_00293",
    pages = "677--694"
}

@article{10.1613/jair.1.12752,
author = {Uma, Alexandra N. and Fornaciari, Tommaso and Hovy, Dirk and Paun, Silviu and Plank, Barbara and Poesio, Massimo},
title = {Learning from Disagreement: A Survey},
year = {2022},
issue_date = {Jan 2022},
publisher = {AI Access Foundation},
address = {El Segundo, CA, USA},
volume = {72},
issn = {1076-9757},
url = {https://doi.org/10.1613/jair.1.12752},
doi = {10.1613/jair.1.12752},
journal = {J. Artif. Int. Res.},
month = jan,
pages = {1385–1470},
numpages = {86}
}

@misc{dutta2023modelingsubjectivitybymimicking,
      title={Modeling subjectivity (by Mimicking Annotator Annotation) in toxic comment identification across diverse communities}, 
      author={Senjuti Dutta and Sid Mittal and Sherol Chen and Deepak Ramachandran and Ravi Rajakumar and Ian Kivlichan and Sunny Mak and Alena Butryna and Praveen Paritosh},
      year={2023},
      eprint={2311.00203},
      archivePrefix={arXiv},
      primaryClass={cs.AI},
      url={https://arxiv.org/abs/2311.00203}, 
}

@inproceedings{prabhakaran-etal-2024-grasp,
    title = "{GRASP}: A Disagreement Analysis Framework to Assess Group Associations in Perspectives",
    author = "Prabhakaran, Vinodkumar  and
      Homan, Christopher  and
      Aroyo, Lora  and
      Mostafazadeh Davani, Aida  and
      Parrish, Alicia  and
      Taylor, Alex  and
      Diaz, Mark  and
      Wang, Ding  and
      Serapio-Garc{\'i}a, Gregory",
    editor = "Duh, Kevin  and
      Gomez, Helena  and
      Bethard, Steven",
    booktitle = "Proceedings of the 2024 Conference of the North American Chapter of the Association for Computational Linguistics: Human Language Technologies (Volume 1: Long Papers)",
    month = jun,
    year = "2024",
    address = "Mexico City, Mexico",
    publisher = "Association for Computational Linguistics",
    url = "https://aclanthology.org/2024.naacl-long.190/",
    doi = "10.18653/v1/2024.naacl-long.190",
    pages = "3473--3492"
}

@article{Kendall1938,
  journal = {Biometrika},
  author = {Kendall, Maurice G.},
  title = {A new measure of rank correlation},
  year = {1938},
  volume = {30},
  number = {1/2},
  pages = {81--93},
  numpages = {13},
  publisher = {[Oxford University Press, Biometrika Trust]},
  issn = {00063444},
  doi = {10.2307/2332226},
}

@misc{piot2025bridginggapshatespeech,
      title={Bridging Gaps in Hate Speech Detection: Meta-Collections and Benchmarks for Low-Resource Iberian Languages}, 
      author={Paloma Piot and José Ramom Pichel Campos and Javier Parapar},
      year={2025},
      eprint={2510.11167},
      archivePrefix={arXiv},
      primaryClass={cs.CL},
      url={https://arxiv.org/abs/2510.11167}, 
}

@article{Poletto2020,
  title = {Resources and benchmark corpora for hate speech detection: a systematic review},
  volume = {55},
  ISSN = {1574-0218},
  url = {http://dx.doi.org/10.1007/s10579-020-09502-8},
  DOI = {10.1007/s10579-020-09502-8},
  number = {2},
  journal = {Language Resources and Evaluation},
  publisher = {Springer Science and Business Media LLC},
  author = {Poletto,  Fabio and Basile,  Valerio and Sanguinetti,  Manuela and Bosco,  Cristina and Patti,  Viviana},
  year = {2020},
  month = sep,
  pages = {477–523}
}

@article{Vidgen2020,
  title = {Directions in abusive language training data,  a systematic review: Garbage in,  garbage out},
  volume = {15},
  ISSN = {1932-6203},
  url = {http://dx.doi.org/10.1371/journal.pone.0243300},
  DOI = {10.1371/journal.pone.0243300},
  number = {12},
  journal = {PLOS ONE},
  publisher = {Public Library of Science (PLoS)},
  author = {Vidgen,  Bertie and Derczynski,  Leon},
  editor = {Grabar,  Natalia},
  year = {2020},
  month = dec,
  pages = {e0243300}
}

@article{Hayes2007,
  title = {Answering the Call for a Standard Reliability Measure for Coding Data},
  volume = {1},
  ISSN = {1931-2466},
  url = {http://dx.doi.org/10.1080/19312450709336664},
  DOI = {10.1080/19312450709336664},
  number = {1},
  journal = {Communication Methods and Measures},
  publisher = {Informa UK Limited},
  author = {Hayes,  Andrew F. and Krippendorff,  Klaus},
  year = {2007},
  month = apr,
  pages = {77–89}
}

\end{document}